\documentclass[11pt,letterpaper]{article}
\usepackage{naaclhlt2015}
\usepackage{times}
\usepackage{latexsym}
\usepackage{amsmath,amsthm,amssymb}
\usepackage{multicol}
\usepackage{graphicx}
\usepackage{float}
\usepackage{caption}
\usepackage{graphicx}
\usepackage{booktabs}
\usepackage[usenames,dvipsnames]{xcolor}
\usepackage{qtree}
\usepackage{hyperref}
\usepackage{subcaption}

\usepackage[round]{natbib}
\renewcommand\newcite[1]{\citet{#1}}
\renewcommand\cite[1]{\citep{#1}}

\newcommand{\ww}{\mathbf{w}}
\newcommand{\uu}{\mathbf{u}}
\newcommand{\sS}{\mathbf{S}}
\newcommand{\mD}{\mathcal{D}}
\newcommand{\vtheta}{\pmb{\theta}}
\title{Structured Prediction of Sequences and Trees using Infinite Contexts}

\author{
\begin{tabular}[t]{cccc}
Ehsan Shareghi\textsuperscript{1}&Gholamreza Haffari\textsuperscript{1}&Trevor Cohn\textsuperscript{2}&Ann Nicholson\textsuperscript{1}
\end{tabular}\\
\begin{tabular}[t]{c}
\textsuperscript{1}Monash University, \textsuperscript{2}University of Melbourne\\
\textsuperscript{1,2}Melbourne, Australia\\
\textsuperscript{1}\texttt{\{ehsan.shareghi,gholamreza.haffari,ann.nicholson\}@monash.edu}\\
\textsuperscript{2}\texttt{t.cohn@unimelb.edu.au}
\end{tabular}
}

\date{}

\begin{document}
\maketitle
\begin{abstract}
Linguistic structures exhibit a rich array of global phenomena, however commonly used Markov models are unable to adequately describe these phenomena due to their strong locality assumptions.  
We propose a novel hierarchical model for structured prediction over sequences and trees which exploits global context by conditioning each generation decision on an \emph{unbounded} context of prior decisions. This builds on the success of Markov models but without imposing a fixed bound in order to better represent global phenomena.
To facilitate learning of this large and unbounded model, we use a hierarchical Pitman-Yor process prior which provides a recursive form of smoothing. We propose prediction algorithms based on A* and  Markov Chain Monte Carlo sampling. 
Empirical results demonstrate the potential of our model compared to baseline finite-context Markov models on part-of-speech tagging and syntactic parsing.
\end{abstract}
%%%%%%%%%%%%%%%%%%%%%%%%%%%%%%%%%%%%%%%%%%%%%%%%%%%%%%
\section{Introduction}

Markov models are widespread popular techniques for modelling the underlying structure of natural language, e.g., as sequences and trees.
However local Markov assumptions often fail to capture phenomena outside the local Markov context, i.e., when the data generation process exhibits long range dependencies.
A prime example is language modelling where only short range dependencies are captured by finite-order (i.e. $n$-gram) Markov models.
%
%An example of such a process is language modeling, where the probability of a word following a sequence of words depends on the previously seen words in its context. Capturing the short range dependencies in language modeling is usually done via finite-order (i.e. n-grams) Markov models. 
However, it has been shown that going beyond finite order in a Markov model  improves language modelling because natural language embodies a large array of long range depepndencies ~\cite{DBLP:conf/icml/WoodAGJT09}. 
While \emph{infinite} order Markov models have been extensively explored for language modelling ~\cite{gasthaus2010improvements,wood2011sequence},
this has not yet been done for structure prediction. %predicting latent structures, such as trees and sequences.

%While the idea of infinite-order Markov models for language modeling task, known as Sequence Memoizer, has been successfully applied ~\cite{gasthaus2010improvements,wood2011sequence},  infinite Markov models for predicting latent trees and sequences has been less explored. This is mainly due to the challenges involved in learning and inference phases.}

In this paper, we propose an infinite-order Markov model for predicting latent structures, namely tag sequences and trees. 
%modeling tree structured data, and sequences with latent structures. 
We show that this expressive model can be applied to various structure prediction tasks in NLP,  such as syntactic parsing and part-of-speech tagging. We propose effective algorithms to tackle  significant learning and inference challenges posed by the infinite Markov model. 
%in ting significant new algorithmic challenges for efficient and accurate inference.

More specifically, we propose an unbounded-depth, hierarchical, Bayesian non-parametric model for the generation of linguistic utterances and their corresponding structure (e.g., the sequence of POS tags or syntax trees). Our model conditions each decision in a tree generating process %, and transitions and emissions in a sequence of POS tags 
on an \emph{unbounded} context consisting of the vertical chain of their ancestors, in the same way that infinite sequence models (e.g., $\infty$-gram language models) condition on an unbounded window of linear context \cite{DBLP:conf/nips/MochihashiS07,wood2009stochastic}.

Learning in this model is particularly challenging due to the large space of contexts and corresponding data sparsity. 
For this reason predictive distributions associated with contexts are smoothed using distribtions for successively smaller contexts via a hierarchical Pitman-Yor process, organised as a trie.
The infinite context  makes it impossible to directly apply dynamic programing for structure prediction.
We present two inference algorithms based on A* and Markov Chain Monte Carlo (MCMC) for predicting the best structure for a given input utterance.

The experiments show that our generative model obtains similar performance to the state-of-the-art Stanford part-of-speech-tagger \cite{toutanova2000enriching} for English and Swedish. For Danish, our model outperforms the Stanford tagger, which is impressive given the Stanford parser uses many more complex features and a discriminative training objective. 
Our experiments on parsing show that our unbounded-context tree model  adapts itself to the data to effectively capture sufficient context to outperform both a PCFG baseline as well as Markov models with finite ancestor conditioning.

%%%%%%%%%%%%%%%%%%%%%%%%%%%%%%%%%%%%%%%%%%%%%%%%%%%%%%
\section{Background and related work}
The syntactic parse tree of an utterance can be generated by combining a set of rules from a grammar, such as a context free grammar (CFG). 
A CFG is a 4-tuple $\mathcal{G}= (\mathcal{T},\mathcal{N},S,\mathcal{R})$, where $\mathcal{T}$ is a set of terminal symbols, $\mathcal{N}$ is a set of non-terminal symbols, $S \in \mathcal{N}$ is the distinguished root non-terminal and $\mathcal{R}$ is a set of productions (a.k.a., rewriting rules). A PCFG assigns a probability to each rule in the grammar, where $\sum_{B,C} P(A\rightarrow B\ C|A) = 1$.
The grammar rules are often in Chomsky Normal Form, taking either the form $A \rightarrow B \ C$ or $A \rightarrow a$ where $A,B,C$ are syntactic cagegories (nonterminals), and $a$ is a word (terminal).

Tag sequences can also be represented as a tree structure, without loss of generality, in which rules take the form $A\rightarrow B\ a$ or $A\rightarrow a$ where $A,B$ are POS tags, and $a$ is a word. Hence tagging models can be represented by restricted (P)CFGs. This unifies view to  syntactic parsing and POS tagging  will allow us to apply our model and inference algorithms to these problems with only minor refinements (see Figure~\ref{fig:parsinghmm}).

In PCFG,  a tree is generated  by starting with the root symbol and rewriting (substituting) it with a grammar rule, then continuing to rewrite frontier non-terminals with grammar rules until there are no remaining frontier non-terminals. When making the decision about the next rule to expand a frontier non-terminal, the only conditioning context used from the partially generated tree is the frontier non-terminal itself, i.e., the rewrite rule is assumed independent from the remainder of the tree given the frontier non-terminal.
Our model relaxes this strong independence assumptions by considering unbounded vertical history when making the next inference decision. This takes into account a wider context when making the next parsing decision.

Perhaps the most relevant work is on  unbounded history language models \cite{DBLP:conf/nips/MochihashiS07,DBLP:conf/icml/WoodAGJT09}. A prime work is Sequence Memoizer  \cite{wood2011sequence} which conditions the generation of the next word on an unbounded history of  previously generated words. 
%However such unbounded models are only applicable to sequence data, while in this paper 
We build on these techniques to develop rich infinite-context models for structured prediction, leading to additional complexity and challenges.

For syntactic parsing, several infinite extensions of probabilistic context free grammars (PCFGs) have been proposed~\cite{liang-EtAl:2007:EMNLP-CoNLL2007,fgm07}. These approaches achieve infinite grammars by allowing an unbounded set of non-terminals (hence grammar rules), but still make use of a bounded history when expanding each non-terminal. An alternative method allows for infinite grammars by considering segmentation of trees into arbitrarily large tree fragments, although only a limited history is used to conjoin fragments \cite{cohn2010inducing,DBLP:conf/nips/JohnsonGG06}. Our work achieves infinite grammars by growing the \emph{vertical} history needed to make the next parsing decision, as opposed to growing the number of rules, non-terminals or states \emph{horizontally}, as done in prior work.

Earlier work in syntactic parsing has also looked into growing both the history vertically and the rules horizontally, in a \emph{bounded} setting. \cite{Johnson:1998:PML:972764.972768} has increased the history for the parsing task by parent-annotation, i.e., annotating each non-terminal in the training parse trees by its parent, and then reading off the grammar rules from the resulting trees. \cite{klein2003accurate} have considered vertical and horizontal markovization while using the head words' part-of-speech tag, and showed that increasing the size of the vertical contexts consistently improves the parsing performance.
%, while bounded horizontal contexts are sufficient.
\cite{petrov2006learning}, \cite{Petrov-Klein-2007:AAAI} and \cite{Matsuzaki:2005:PCL:1219840.1219850} have treated non-terminal annotations as latent variables and estimated them from the data.  

\noindent Likewise, finite-state hidden Markov models (HMMs) have been extended \emph{horizontally} to have countably infinite number of states \cite{DBLP:conf/nips/BealGR01}. Previous works on applying Markov models to part-of-speech tagging  either considered finite-order Markov models ~\cite{DBLP:conf/anlp/Brants00}, or finite-order HMM~\cite{Thede:1999:SHM:1034678.1034712}. We differ from these works by 
conditioning \emph{both} the emissions and transitions on their \emph{full} contexts.
%%%%%%%%%%%%%%%%%%%%%%%%%%%%%%%%%%%%%%%%%%%%%%%%%%%%%%%%%%%%%
\section{The Model}
\begin{figure}
\footnotesize
  \centering
\includegraphics[width=1\columnwidth]{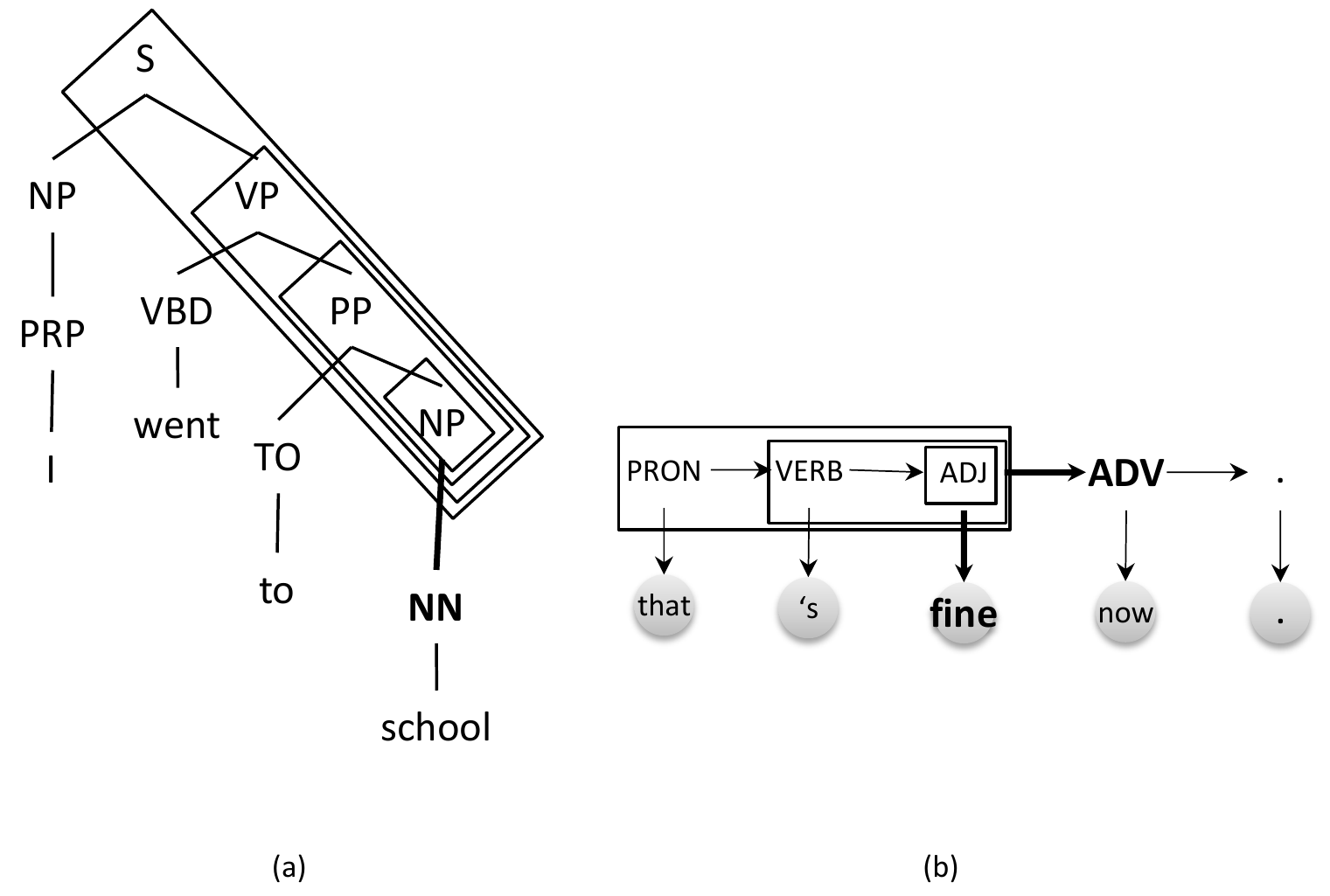}
\caption{Examples of infinite-order conditioning and smoothing mechanism. The bold symbols (\textbf{NN, ADV, fine}) are the part of the structure being generated, and the boxes correspond to the conditioning context. (a) Syntactic Parsing, and (b) Infinite-order HMM for POS tagging.}
\label{fig:parsinghmm}
\end{figure}

Our model relaxes strong local Markov assumptions in PCFG to enable capturing phenomena outside of the local Markov context.
The model conditions the generation of a rule in a tree on its unbounded  vertical history, i.e.,  its ancestors on the path towards the root of the tree (see Figure \ref{fig:parsinghmm}). Thus the probability of a tree $T$ is
$$ P(T) = \prod_{(\uu,r) \in T} G_{[\uu]}(r)$$
where $r$ denotes the rule and  $\uu$  its history, and $G_{[\uu]}(.)$ is the probability of the next inference decision (i.e., grammar rule) conditioned on the context $\uu$.
In other words, a tree $T$ can be represented as a sequence of context-rule events $\{(\uu,r) \in T\}$.

When learning such a model from data, a vector of predictive probabilities for the next rule $G_{[\uu]}(.)$ given each possible vertical context $\uu \in \mathcal{U}$ must be learned, where  depending on the problem $\mathcal{U}$ can denote the set of chains of non-terminals $\mathcal{N}^*$ or chains of rules $\mathcal{R}^*$.
As the context size increases, the number of events observed for such long contexts in the training data  drastically decreases which makes parameter estimation challenging, particularly when generalising to unseen contexts.  
Assuming our unbounded-depth model, we need suitable \emph{smoothing} techniques to estimate conditional rule probabilities for large (and possibly infinite depth) contexts. 
We achieve smoothing  by placing a hierarchical Bayesian prior over the set of probability distributions $\{G_{[\uu]}\}_{u \in \mathcal{U}}$. 
We smooth $G_{[\uu]}$ with a distribution conditioned on a shorter context  $G_{[\pi(\uu)]}$, where $\pi(\uu)$ is the suffix of $\uu$ containing all but the earliest event.
This ties  parameters of  longer histories to their shorter suffixes in a hierarchical manner, and leads to sharing statistical strengths to overcome sparsity issues. Figure~\ref{fig:parsinghmm} shows our infinite-order Markov model and the smoothing mechanism described here.

\begin{figure}
\footnotesize
  \centering
\includegraphics[width=0.45\textwidth]{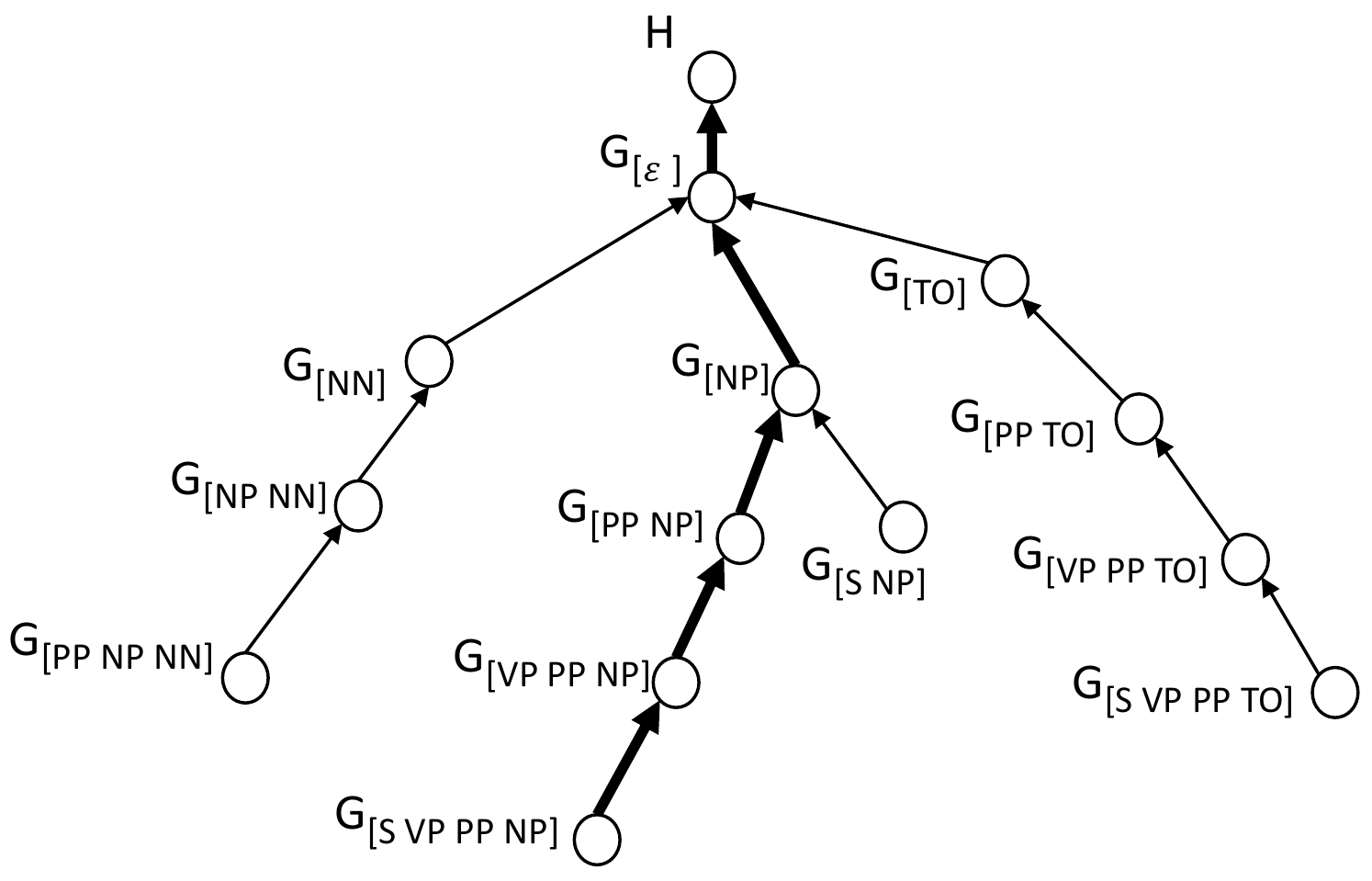}
\caption{Part of the smoothing mechanism corresponding to Figure 1(a). Each node represents a distribution $G$ labeled with a context, and the directed edges demonstrate the direction of smoothing. The path in bold corresponds to the smoothing for the \emph{rule} $NP\rightarrow NN$.}
\label{fig:hpyp}
\end{figure}

More specifically, we assume that a distribution with the full history $G_{[u]}$ is related to a distribution with the most recent history $G_{[\pi(u)]}$ through the Pitman-Yor process $PYP$ \cite{wood2011sequence}: 
\begin{align*}
G_{[\varepsilon]}~|\ d_{[\varepsilon]},c_{[\varepsilon]},H\ &\sim\  PYP(d_{0},c_{0},H)\\
G_{[\uu]}~|\ d_{|\uu|},c_{|\uu|},G_{[\pi(\uu)]}\ &\sim\  PYP(d_{|\uu|},c_{|\uu|},G_{[\pi(\uu)]})
\end{align*}
where $H$ denotes the  base (e.g.  uniform) distribution, and $\varepsilon$ denotes the empty context. 
The Pitman-Yor process $PYP(d,c,H)$ is a distribution over distributions, where $d$ is the discount parameter, $c$ is the concentration parameter, and H is the base distribution.  
Note that $G_{[u]}$ depends on $G_{[\pi(u)]}$ which itself  depends on $G_{[\pi(\pi(u))]}$, etc. This leads to a hierarchical Pitman-Yor process prior where context-dependent distributions are \emph{hidden}.
The formulation of the hierarchical PYP over different length contexts is illustrated in Figure~\ref{fig:hpyp}.

\begin{figure}
\centering
\begin{subfigure}{0.65\columnwidth}
  \centering
  \includegraphics[width=1\textwidth]{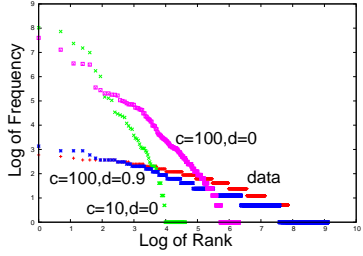}
  \caption{$\uu:$ S NP}
\end{subfigure}%

\begin{subfigure}{0.65\columnwidth}
  \centering
  \includegraphics[width=1\textwidth]{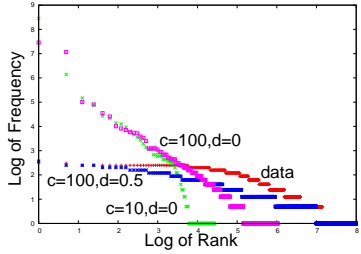}
  \caption{$\uu:$ VERB}
\end{subfigure}
\caption{$\log$-$\log$ plot of rule frequency vs rank, illustrated for (a) syntactic parsing and (b) POS tagging. Besides the data distribution, we also show samples from three PYP distributions with different hyperparameter values, $c,d$.}
\label{pypproperty}
\end{figure}

Figure~\ref{pypproperty} demonstrates the property of PYP and how its behaviour depends on discount $d$, and concentration $c$ parameters.
Note that the PYP allows a good fit to data distribution compared to the Dirichlet Process ($d=0$; as used in prior work) which cannot adequately represent the long tail of events.

%%%%%%%%%%%%%%%%%%%%%%%%%%%%%%%%%%%%%%%%%%%%%%%%%%%%%%
\section{Learning}
Given a training tree-bank, i.e., a collection of utterances and their trees, we are interested in the posterior distribution over $\{G_{[\uu]}\}_{\uu \in \mathcal{U}}$.
We make use of the approach developed in \newcite{wood2011sequence} for learning such suffix-based graphical models when learning infinite-depth language models. It makes use of Chinese Restaurant Process (CRP) representation of the Pitman-Yor process in order to marginalize out distributions $G_{[\uu]}$ \cite{teh2006hierarchical} and learn the predictive probabilities $P(r | \uu)$. 

Under the CRP representation each context corresponds to a restaurant. As a new $(\uu,r)$ is observed in the training data, a \emph{customer} is entered to the restaurant, i.e., the trie node corresponding to $\uu$. 
Whenever a customer enters a restaurant, it should be decided whether to seat him on an existing table serving the \emph{dish} $r$, or to seat him on a new table  and sending a proxy customer to the parent node in the trie to order $r$  (i.e., based on $(\pi(\uu),r)$). 
Fixing a seating arrangement $\sS$ and PYP parameters ${\vtheta}$ for all restaurants (i.e., the collection of concentration and discount parameters), the predictive probability of a rule based on our infinite-context rule model is:
%{\footnotesize
\begin{align*}
P(r|\epsilon,\sS,\vtheta) &= H(r)\\
P(r|\uu,\sS,\vtheta) &=
\frac{n^{\uu}_{r.}-d_{|\uu|}t^{\uu}_{r}}{n^{|\uu|}_{..}+c_{|\uu|}} \\
&+\frac{c_{|\uu|}+d_{|\uu|}t^{\uu}_{.}}{n^{\uu}_{..}+c_{|\uu|}}P(r|\pi(\uu),\sS,\vtheta)
\end{align*}
%}
where $d_{|\uu|}$ and $c_{|\uu|}$ are the discount and concentration parameters, $n^{\uu}_{rk}$ is the number of customers at table $k$ served the dish $r$ in the restaurant $\uu$ (accordingly $n^{\uu}_{r.}$ is the number of customers served the dish $r$ and $n^{\uu}_{..}$ is the  number of customers),  
%($c^{u}_{rk} = 0$ if the $k$th table does not serve dish $r$), 
and $t^{\uu}_r$ is the number of tables serving dish $r$ in the restaurant  $\uu$ (accordingly $t^{\uu}_.$ is the  number of tables).

The seating arrangements (the state of all restaurants including their tables and customers sitting on each table) are hidden, so they need to be marginalized out:
\begin{equation*}
P(r|\uu,\mD) = \int P(r|\uu,\sS,\vtheta)P(\sS,\vtheta |\mD)d(\sS,\vtheta)
\end{equation*}
where $\mD$ is the training tree-bank. 
We approximate this integral by the so called ``minimal assumption seating arrangement''  and the MAP parameter setting $\vtheta$ which maximizes the corresponding data posterior. 
Based on  the minimal assumption, a new table is created only when there is no table serving the desired dish in a restaurant $\uu$. That is, a proxy customer is created and sent to the parent node in the trie  $\pi(\uu)$ for each unique dish type (sequence of events). 

This approximation has been shown to recover interpolated Kneser-Ney smoothing, when  applied to hierarchical Pitman-Yor process  language model \cite{teh2006hierarchical}.

The parameter $\vtheta$ is learned by maximising the posterior, given the seating arrangement corresponding to the minimal assumption.
We put the following prior distributions over the parameters: $d_m \sim \textrm{Beta}(a_m, b_m)$ and $c_m \sim \textrm{Gamma}(\alpha_m,\beta_m)$. The posterior is the prior multiplied by the following likelihood term:
\begin{align*}
\prod_r H(r)^{n^0_{r.}} \prod_{\uu} \frac{[c_{|\uu|}]^{t^{\uu}_.}_{d_{|\uu|}}}{[c_{|\uu|}]_1^{n^{\uu}_{..}}} \prod_r \prod_{k=1}^{t^{\uu}_.} [1 - d_{|\uu|}]_1^{(n^{\uu}_{rk}-1)}
\end{align*}
where $[a]_b^{c}$ denotes the generalised factorial function.\footnote{$[a]^{0}_b = [a]^{-1}_b = 1$ and $[a]^b_c = \prod_{i=0}^{c-1} (a + i b)$.} %
%Combining the prior with likelihood, we maximise the posterior distribution, with the constraints $c_m > - d_m$ and $d_m \in [0,1)$. 
We maximize the posterior with the constraints $c_m \ge 0$ and $d_m \in [0,1)$ using the L-BFGS-B optimisation method \cite{zhu1997algorithm}, which results in  the optimised discount and concentration values for each context size.

%%%%%%%%%%%%%%%%%%%%%%%%%%%%%%%%%%%%%%%%%%%%%%%%%%%%%%
\section{Prediction}

In this section, we propose  algorithms for the challenging problem of predicting the highest scoring tree. 
%Our algorithms are \emph{top-down}, hence have access to the full history when deciding about the next parsing decision.   
The key ideas are to compactly represent the \emph{space} of all possible trees for a given utterance, and then \emph{search} for the best tree in this space in a \emph{top-down} manner. 
By  traversing the hyper-graph top-down, the search algorithms  have access to the full history of grammar rules.

In the test time, we  need to predict the tree structure of a given utterance $\ww$ by maximizing the tree score:
\begin{eqnarray*} 
\arg\max_T P(T|\mathcal{D},\ww) = \arg\max_T \prod_{(\uu,r) \in T} P(r|\uu,\mathcal{D})
\end{eqnarray*}
The unbounded context allowed by our model makes it infeasible to apply dynamic programming, e.g. CYK \cite{cocke1970programming}, for finding the highest scoring tree.  CYK is a \emph{bottom-up} algorithm which requires storing in a dynamic programming table the score of each utterance's sub-span  conditioned on all possible contexts. Even truncating the context size to bound this term may be insufficient to allow CYK  for prediction, due to the unreasonable computational complexity.

\begin{figure}\footnotesize
  \centering
\includegraphics[width=1\columnwidth]{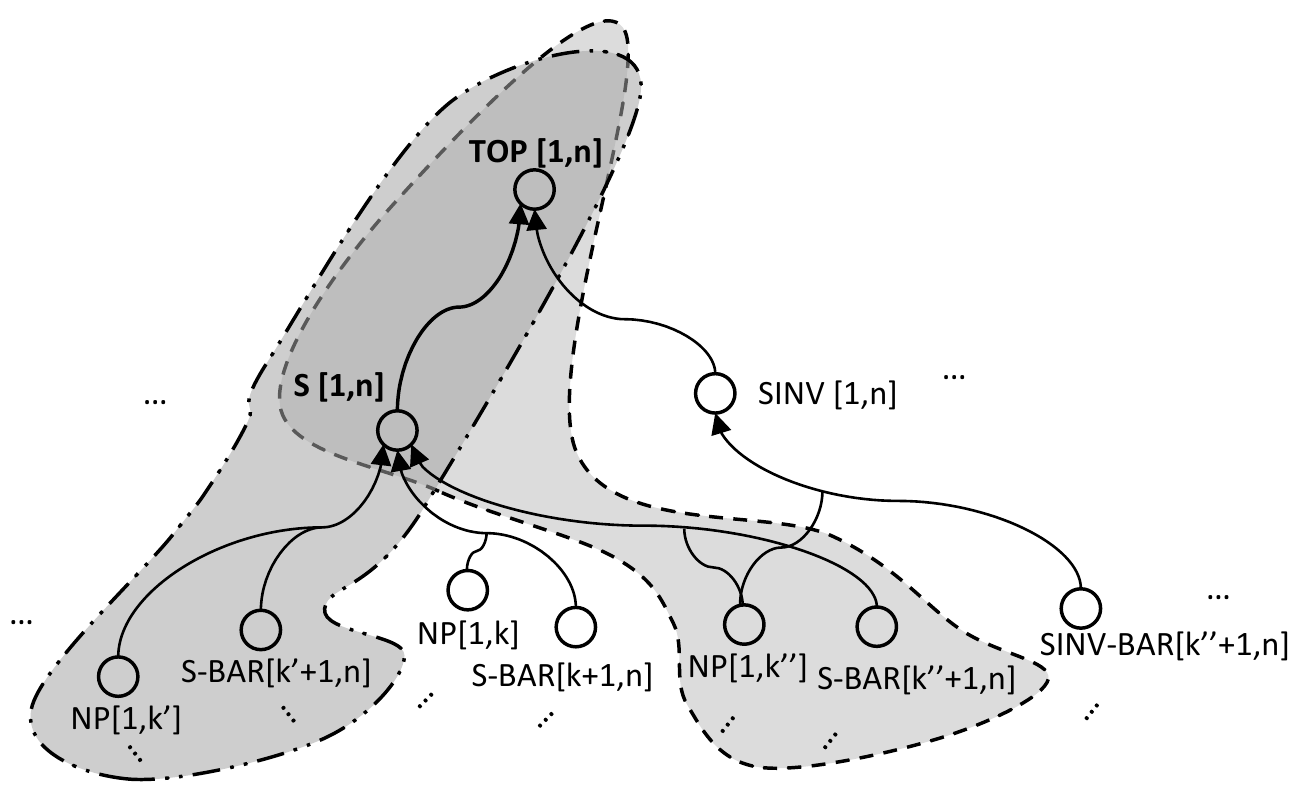}
\caption{{Hyper-graph representation of the search space. The gray areas are examples of two partial hypotheses in A* priority queue.}}
\label{fig:hyperhgraph}
\end{figure}

The space of all possible trees for a given utterance can be compactly represented as a \emph{hyper-graph} \cite{DBLP:conf/iwpt/KleinM01}. Each hyper-graph node is labelled with a non-terminal and a sub-span of the utterance. There exists a hyper-edge from the nodes $B[i,j]$ and $C[j+1,k]$ to the node $A[i,k]$ if the rule $A \rightarrow B \ C$ belongs to the grammar (Figure \ref{fig:hyperhgraph}). Starting from the top node $S[0,N]$, our prediction algorithms  search for the highest scoring tree sub-graph that covers all of the utterance terminals in  the  hyper-graph. 
Our top-down prediction algorithms  have access to the full history needed by our model when deciding about the next hyper-edge to be added to the  partial tree.

%%%%%%%%%%%%%%%%%%%%%%%%%%%%%%%%%%%%%%%%%%%%%%%%%%%%%%
\subsection{A* Search}

This algorithm incrementally expands frontier nodes of the best partial tree until a complete tree is constructed.  In the expansion step, all possible rules for expanding all frontier non-terminals are considered and the resulting partial trees are inserted into a priority queue (see Figure~\ref{fig:hyperhgraph}), sorted based on the following score:
\begin{eqnarray*} 
\label{score}
Score(T^+) &=& \log P(T) + \log G_{\uu}(A \rightarrow B \ C) \\
         &+& h(T^+, A \rightarrow B \ C,i,k,j|G')
\end{eqnarray*}
where $T^+$ is a partial tree \emph{after} expanding a frontier non-terminal, $P(T)$ is the probability of the current partial tree, $G_{\uu}(A \rightarrow B \ C)$ is the probability of expanding  a non-terminal via a rule $A \rightarrow B \ C$ in the full context $\uu$, and $h$ is the heuristic function (i.e., the estimate of the score for the best tree completing $T^+$). We use various heuristic functions when expanding a node $A[i,j]$ in the hypergraph via a hyperedge with tails $B[i,k]$ and $C[k+1,j]$: 
\begin{itemize}
\item \textbf{Full Frontier:} which estimates the completion cost by 
\begin{eqnarray*} 
&&h(T^+, A \rightarrow B \ C,i,k,j|G') = \\
&& \sum_{(A',i',j') \in \textrm{Fr}(T^+)} \log P(A',i',j'|G')
\end{eqnarray*}
%
%$$h(T^+, A \rightarrow B \ C,i,k,j|G') = \sum_{(A',i',j') \in \textrm{Fr}(T^+)} \log P(A',i',j'|G')$$ 
where $\textrm{Fr}(T^+)$ is the set of frontier nodes of the partial tree, and $G'$ is a \emph{simplified grammar} admitting dynamic programming. Here we choose the PCFG used the base measure $H$ in the root of the PYP hierarchy. Accordingly the $\log P$ terms can be computed cheaply using the PCFG inside probabilities.
\item \textbf{Local Frontier:} which only takes into account the completion of the following frontier nodes:
\begin{eqnarray*}
&&h(T^+, A \rightarrow B \ C,i,k,j|G')= \\
&&\log P(B,i,k|G') + \log P(C,k+1,j|G')
\end{eqnarray*}
This heuristic focuses on the completion cost of the sub-span using the selected rule.
\end{itemize}
The above heuristics functions are not admissible, hence the A* algorithm  is not guaranteed to find the optimal  tree. However the PCFG provides reasonable estimates of the completion costs, and accordingly with a sufficiently wide beam, search error is likely to be low.

\subsection{MCMC Sampling}

We make use of Metropolis-Hastings (MH) algorithm, which is a Markov chain Monte Carlo (MCMC) method, for obtaining a sequence of random trees. We then combine these trees to construct the predicted tree.

In the MH algorithm, we use a \textit{PCFG} as our \emph{proposal} distribution $Q$ and draw samples from it. Each sampled tree is then accepted/rejected using the following acceptance rate:
\begin{eqnarray*} 
\alpha(T,T') = \min\left\{1,\frac{P(T')Q(T)}{P(T)Q(T')}\right\}
\end{eqnarray*}
where $T'$ is the sampled tree, $T$ is the current tree, $P(T')$ is the probability of the proposed tree under our model, and $Q(T')$ is its probability under the proposal PCFG. Under some conditions, i.e., detailed balance and ergodicity, it is guarantheed that the stationary distribution of the underlying Markov chain (defined by the MH sampling) is the distribution that our model induces over the space of trees $P$. 
For each utterence, we sample a fresh tree for the whole utterance from a PCFG using the approach of \cite{johnson2007bayesian}, which works by first computing the inside lattice under the proposal model (which can be computed once and reused), followed by top-down sampling to recover a tree. Finally the proposed tree is scored using the MH test, according to which the tree is randomly accepted as the next sample or else rejected in which case the previous sample is retained.

Once the sampling is finished, we need to choose a tree based on statistics of the sampled collection of trees. One approach is to select the most frequently sampled tree, however this does not work effectively in such large search spaces because of high sampling variance. Note that local Gibbs samplers might be able to address this problem, at least partly, through resampling subtrees instead of full tree sampling (as done here). Local changes would allow for more rapid mixing from trees with some high and low scoring subtrees to trees with uniformly high scoring sub-structures. 
We leave local sampling for future work, noting that the obvious local operation of resampling complete sub-trees or local tree fragments would compromise detailed balance, and thus not constitute a valid MCMC sampler~\cite{levenberg2012bayesian}.

To address this problem, we use a Minimum Bayes Risk (MBR) decoding method to predict the best tree \cite{Goodman:1996:PAM:981863.981887} as follows: For each pair of a nonterminal-span, we record the count in the collection of sampled trees. Then using the Viterbi algorithm, we select the tree from the hypergraph for which the sum of the induced pairs of nonterminal-span is maximized. Roughly speaking, this allows to make local corrections that result in higher accuracy compared to the best sampled trees.
%%%%%%%%%%%%%%%%%%%%%%%%%%%%%%%%%%%%%%%%%%%%%%%%%%%%%%
\section{Experiments}
In order to evaluate the proposed model and prediction algorithms, we performed two sets of experiments on tasks with different structural complexity.
%and to compare their performance on datasets with different structural complexity we performed three sets of experiments. 
The statistics of the tasks and datasets are provided in Table \ref{charac}.
 \begin{table}[t]\footnotesize
%\resizebox{\columnwidth}{!}{
\centering
 \begin{tabular}{lcccc}
 Task  & Train & Test & Len & Rules\\\hline
%\textbf{morph} & 36479 & 4000 & 5 & 3080\\
\textbf{parse} & 33180 & 2416 & 24 & 31920\\
\textbf{pos EN} & 38219 & 5462 & 24 & 29499\\
\textbf{pos DN} & 3638 & 1000 & 20 & 5269\\
\textbf{pos SW} & 10653 & 389 & 18 & 9739\\
 \end{tabular}
 %}
\caption{Statistics for PTB syntactic Parsing and part-of-speech tagging, showing the number of training and test sentences, average sentence length in words and number of grammar rules. For morph the numbers are averaged over the 10 folds.
%The numbers for the word segmentation datasets are the average between 10-fold splits. 
%The numbers reported for Penn. and wseg are based on the binarized datasets.
}
\label{charac}
\end{table}
%%%%%%%%%%%%%%%%%%%%%%%%%%%%%%%%%%%%%%%%%%%%%%%%%%%%%%
\subsection{Syntactic Parsing} 
For syntactic parsing, we use the  
Penn.~treebank (PTB) dataset~\cite{Marcus1993building}.
%in which  each sentence is annotated with its syntactic structure. 
We used the standard data splits for training and testing (train sec 2-21; validation sec 22; test sec 23). We followed \newcite{petrov2006learning} preprocessing steps by right-binarizing the trees and replacing words with $count\leq1$ in the training sample with generic \emph{unknown} word markers representing the tokens' lexical features and position. The results reported in Table~\ref{result} are produced by EVALB.
%%%%%%%%%%%%%%%%%%%%%%%%%%%%%%%%%%%%%%%%%%%%%%%%%%%%%%

%\subsubsection{Discussion}
The results  in Table~\ref{result} demonstrate the superiority of our model compared to the baseline PCFG.
We note that the A* parser becomes less effective (even with a large beam size) for this task, which we attribute to the large search arising for the large grammar and  long sentences.
%
%of the search space considering the magnitude of the space of possible trees for any given sentence. 
%
%The size of this space is very large given the number of rules (31920), and the average sentence length (24) which then translates into having larger number of possibilities in the search space. 
%
Our best results are achieved by MCMC, demonstrating the effectiveness of MCMC in large search spaces.

An interesting observation is how our results compare with those achieved by bounded vertical and horizontal Markovization reported in~\cite{klein2003accurate}. Our binarization corresponds to one of their simpler settings for horizontal markovization, namely $h=0$ in their terminology, and note also that we ignore the head information which is used in their models. Despite this we still manage to equal their results obtained using vertical context of size 3 ($v=3$), with $76.7$ F1 score. Their best result, $F_1=79.74$, was achieved with $h\leq2$, $v=3$ (and tags for head words). We believe that our model would outperform theirs if we consider greater horizontal markovization and incorporate head word information. To facilitate a fair comparison with vertical markovization, we experimented with limiting the size of the vertical contexts to 2, 3 or 4 within our model. Using  MCMC parsing  we found that performance  consistently improved as the size of the context was increased, scoring $68.1$, $71.1$, $75.0$ F-measure respectively. This is below $76.7$ F-measure of our unbounded-context model which adapts itself to data to effectively capture the right context.

Overall our approach significantly outperforms the baseline PCFG, although note these results are well below the current state-of-the-art in parsing, which typically makes use of discriminative training with much richer features. We speculate that future enhancements  could close the gap between our results and that of modern parsers, while offering the potential benefits of our generative model which allows further incorporation of different types of contexts (e.g., head words and $n$-gram lexical context).
\begin{table}
\resizebox{\columnwidth}{!}{
\centering
 \begin{tabular}{l|lr|lr}
  \multicolumn{1}{c}{}&\multicolumn{2}{c}{all}&\multicolumn{2}{c}{$\leq40$}\\
  \cmidrule(r){2-3} \cmidrule(r){4-5}
  \textbf{Syntactic Parser}&F1&ACC&F1&ACC\\\hline
  A* (Local Frontier)&75.33&16.12&76.21&16.85\\
  A* (Full Frontier)&72.27&13.14&72.34&13.57\\
  MCMC &76.74&18.23&78.21&18.99\\
  %MCMC (STreeSampling)&76.51&17.80&77.32&18.03\\
  \hline
  PCFG CYK&58.91&4.11&60.25&4.42\\
  \end{tabular}}
\caption{Syntactic parsing results for the Penn. treebank, showing labelled F-Measure (F1) and exact bracketing match (ACC).}
\label{result}
\end{table}
%%%%%%%%%%%%%%%%%%%%%%%%%%%%%%%%%%%%%%%%%%%%%%%%%%%%%%
\subsection{Part-of-Speech Tagging}\label{posexperiment}
\begin{figure}
\footnotesize
  \centering
\includegraphics[width=0.45\textwidth]{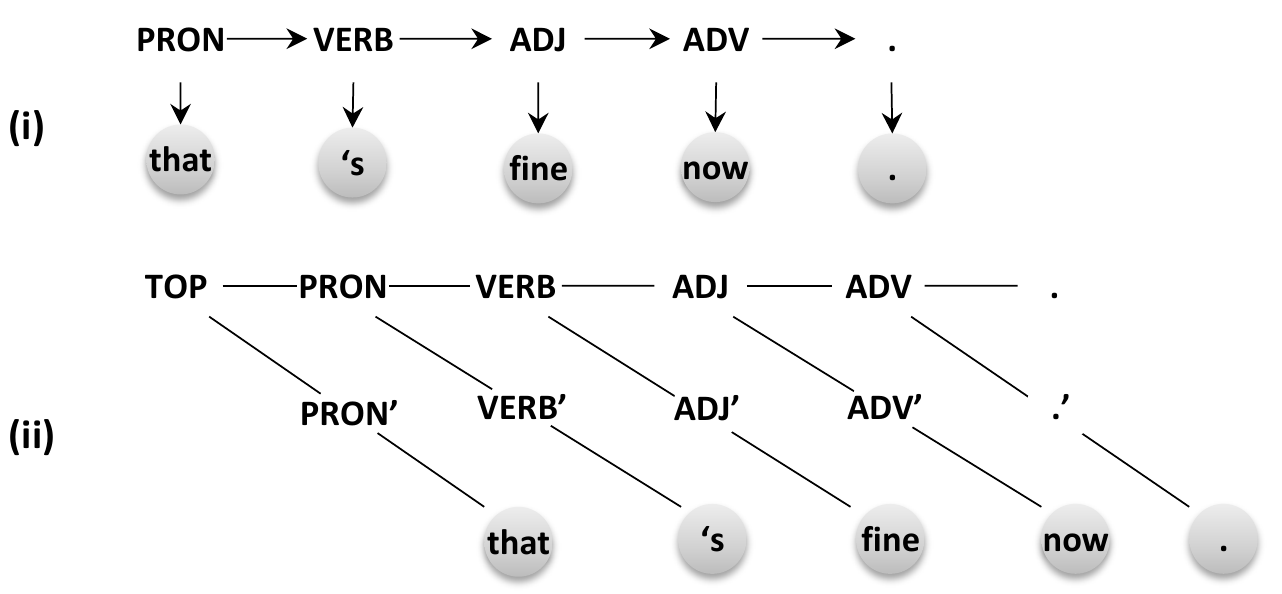}
\caption{The analogy between HMM (i) and our representation (ii) for the part-of-speech tags of the sentence \textit{``that's fine now."}}
\label{fig_pos}
\end{figure}

The part of speech (POS) corpora have been extracted from PTB (sections 0-18 for training and 22-24 for test) for English, and NAACL-HLT 2012 Shared task on Grammar Induction\footnote{\tiny\url{http://wiki.cs.ox.ac.uk/InducingLinguisticStructure/SharedTask}} for Danish and Swedish~\cite{W12-1909}. We convert the sequence of part-of-speech tags for each sentence into a tree structure analogous to a Hidden Markov Model (HMM). For each POS tag we introduce a twin (e.g., ADJ' for ADJ) in order to encode HMM-like transition and emission probabilities in the grammar.  As shown in Figure~\ref{fig_pos}, this representation guarantees that all the rules in the structures are either in the form of $t_i\rightarrow t_j\ t_j'$ (transition) or $t'\rightarrow \texttt{word}$ (emission).

The tagging results are reported in Table~\ref{pos_result}, including  comparison with the baseline PCFG ($\equiv$ HMM) and the state-of-the-art Stanford POS Tagger \cite{toutanova2000enriching}, which we  trained and tested on these datasets.
%%%%%%%%%%%%%%%%%%%%%%%%%%%%%%%%%%%%%%%%%%%%%%%%%%%%%%
%\subsubsection{Discussion}
As illustrated in Table~\ref{pos_result}, our model consistently improves the PCFG baseline. While for Danish we outperform the state-of-the-art tagger, the results for English and Swedish we are a little behind the Stanford Tagger. This is an promising result since our model is only based on the rules and their contexts, as opposed to the Stanford Tagger which uses  complex hand-designed features and a complex form of discriminative training.

Note the strong performance of MCMC sampling, which consistently outperforms A* search on the three tagging tasks.

\begin{table}[t]
\resizebox{\columnwidth}{!}{
 \begin{tabular}{l|lr|lr|lr}
  \multicolumn{1}{c}{}&\multicolumn{2}{c}{\textbf{English}}&\multicolumn{2}{c}{\textbf{Danish}}&\multicolumn{2}{c}{\textbf{Swedish}}\\
  \cmidrule(r){2-3} \cmidrule(r){4-5}\cmidrule(r){6-7}
  \textbf{POS Tagger}&TL&SL&TL&SL&TL&SL\\\hline
  A*(Local Frontier)&95.50&54.11&89.85&35.10&87.04&32.13\\
  A*(Full Frontier)&95.27&53.88&88.57&32.6&85.62&28.53\\
  MCMC &96.04&54.25&95.55&72.93&89.97&34.45\\
  \hline
  PCFG CYK&94.69&47.22&89.04&31.7&89.76&33.93\\\hline
  Stanford Tagger
  &97.24&56.34&93.66&51.30&91.28&37.02\\
 \end{tabular}}
\caption{TL stands for Token-Level Accuracy, SL stands for Sentence-Level Accuracy. MCMC results are the average of 10 runs.}
\label{pos_result}
\end{table}
%%%%%%%%%%%%%%%%%%%%%%%%%%%%%%%%%%%%%%%%%%%%%%%%%%%%%%
\section{Conclusion and Future Work}
We  have proposed a novel hierarchical model over linguistic trees which exploits global context by conditioning the generation of a rule in a tree on an unbounded tree context consisting of the vertical chain of its ancestors.
To  facilitate  learning  of such a large and unbounded model, the predictive distributions associated with tree contexts are smoothed in a recursive manner using a hierarchical Pitman-Yor process.  
We  have shown  how  to  perform  prediction  based  on  our model  to  predict  the  parse  tree  of  a  given   utterance  using  various  search  algorithms,  e.g.  A* and Markov Chain Monte Carlo. 

This consistently improved over baseline methods in two tasks, and produced state-of-the-art results for  Danish part-of-speech tagging.

In future, we would like to consider sampling the seating arrangements and model hyperparameters, and seek to incorporate several different notions of context besides the chain of ancestors.
%%%%%%%%%%%%%%%%%%%%%%%%%%%%%%%%%%%%%%%%%%%%%%%%%%%%%%
%\bibliographystyle{naaclhlt2015} % see comment in header
\bibliographystyle{abbrvnat}
\small
\bibliography{Bibliography}

\begin{thebibliography}{27}
\providecommand{\natexlab}[1]{#1}
\providecommand{\url}[1]{\texttt{#1}}
\expandafter\ifx\csname urlstyle\endcsname\relax
  \providecommand{\doi}[1]{doi: #1}\else
  \providecommand{\doi}{doi: \begingroup \urlstyle{rm}\Url}\fi

\bibitem[Beal et~al.(2001)Beal, Ghahramani, and
  Rasmussen]{DBLP:conf/nips/BealGR01}
M.~J. Beal, Z.~Ghahramani, and C.~E. Rasmussen.
\newblock The infinite hidden markov model.
\newblock In \emph{Advances in Neural Information Processing Systems,
  Vancouver, British Columbia, Canada}, pages 577--584, 2001.

\bibitem[Brants(2000)]{DBLP:conf/anlp/Brants00}
T.~Brants.
\newblock Tnt -- {A} statistical part-of-speech tagger.
\newblock In \emph{Proceedings of the sixth conference on Applied natural
  language processing}, pages 224--231, 2000.

\bibitem[Cocke and Schwartz(1970)]{cocke1970programming}
J.~Cocke and J.~T. Schwartz.
\newblock Programming languages and their compilers : preliminary notes.
\newblock \emph{Technical report}, 1970.

\bibitem[Cohn et~al.(2010)Cohn, Blunsom, and Goldwater]{cohn2010inducing}
T.~Cohn, P.~Blunsom, and S.~Goldwater.
\newblock Inducing tree-substitution grammars.
\newblock \emph{The Journal of Machine Learning Research}, 11:\penalty0
  3053--3096, 2010.

\bibitem[Finkel et~al.(2007)Finkel, Grenager, and Manning]{fgm07}
J.~Finkel, T.~Grenager, and C.~Manning.
\newblock The infinite tree.
\newblock In \emph{Proceedings of the 45th annual meeting of Association for
  Computational Linguistics}, pages 272--279, 2007.

\bibitem[Gasthaus and Teh(2010)]{gasthaus2010improvements}
J.~Gasthaus and Y.~W. Teh.
\newblock Improvements to the sequence memoizer.
\newblock In \emph{Advances in Neural Information Processing Systems}, pages
  685--693, 2010.

\bibitem[Gelling et~al.(2012)Gelling, Cohn, Blunsom, and Graca]{W12-1909}
D.~Gelling, T.~Cohn, P.~Blunsom, and J.~Graca.
\newblock \emph{Proceedings of the NAACL-HLT Workshop on the Induction of
  Linguistic Structure}, chapter The PASCAL Challenge on Grammar Induction,
  pages 64--80.
\newblock Association for Computational Linguistics, 2012.

\bibitem[Goodman(1996)]{Goodman:1996:PAM:981863.981887}
J.~Goodman.
\newblock Parsing algorithms and metrics.
\newblock In \emph{Proceedings of the 34th Annual Meeting on Association for
  Computational Linguistics}, ACL '96, pages 177--183, Stroudsburg, PA, USA,
  1996. Association for Computational Linguistics.

\bibitem[Johnson(1998)]{Johnson:1998:PML:972764.972768}
M.~Johnson.
\newblock Pcfg models of linguistic tree representations.
\newblock \emph{Computational Linguistics}, 24\penalty0 (4):\penalty0 613--632,
  Dec. 1998.
\newblock ISSN 0891-2017.

\bibitem[Johnson et~al.(2006)Johnson, Griffiths, and
  Goldwater]{DBLP:conf/nips/JohnsonGG06}
M.~Johnson, T.~L. Griffiths, and S.~Goldwater.
\newblock Adaptor grammars: {A} framework for specifying compositional
  nonparametric bayesian models.
\newblock In \emph{Advances in Neural Information Processing Systems 19,
  Proceedings of the Twentieth Annual Conference on Neural Information
  Processing Systems, Vancouver, British Columbia, Canada, December 4-7, 2006},
  pages 641--648, 2006.

\bibitem[Johnson et~al.(2007)Johnson, Griffiths, and
  Goldwater]{johnson2007bayesian}
M.~Johnson, T.~L. Griffiths, and S.~Goldwater.
\newblock Bayesian inference for pcfgs via markov chain monte carlo.
\newblock In \emph{HLT-NAACL}, pages 139--146, 2007.

\bibitem[Klein and Manning(2001)]{DBLP:conf/iwpt/KleinM01}
D.~Klein and C.~D. Manning.
\newblock Parsing and hypergraphs.
\newblock In \emph{Proceedings of the Seventh International Workshop on Parsing
  Technologies (IWPT-2001), 17-19 October 2001, Beijing, China}, 2001.

\bibitem[Klein and Manning(2003)]{klein2003accurate}
D.~Klein and C.~D. Manning.
\newblock Accurate unlexicalized parsing.
\newblock In \emph{Proceedings of the 41st Annual Meeting on Association for
  Computational Linguistics-Volume 1}, pages 423--430. Association for
  Computational Linguistics, 2003.

\bibitem[Levenberg et~al.(2012)Levenberg, Dyer, and
  Blunsom]{levenberg2012bayesian}
A.~Levenberg, C.~Dyer, and P.~Blunsom.
\newblock A bayesian model for learning scfgs with discontiguous rules.
\newblock In \emph{Proceedings of the 2012 joint conference on empirical
  methods in natural language processing and computational natural language
  learning}, pages 223--232. Association for Computational Linguistics, 2012.

\bibitem[Liang et~al.(2007)Liang, Petrov, Jordan, and
  Klein]{liang-EtAl:2007:EMNLP-CoNLL2007}
P.~Liang, S.~Petrov, M.~Jordan, and D.~Klein.
\newblock The infinite {PCFG} using hierarchical {Dirichlet} processes.
\newblock In \emph{Proceedings of the 2007 Joint Conference on Empirical
  Methods in Natural Language Processing and Computational Natural Language
  Learning (EMNLP-CoNLL)}, pages 688--697, 2007.

\bibitem[Marcus et~al.(1993)Marcus, Marcinkiewicz, and
  Santorini]{Marcus1993building}
M.~P. Marcus, M.~A. Marcinkiewicz, and B.~Santorini.
\newblock Building a large annotated corpus of english: The penn treebank.
\newblock \emph{Computational linguistics}, 19\penalty0 (2):\penalty0 313--330,
  1993.

\bibitem[Matsuzaki et~al.(2005)Matsuzaki, Miyao, and
  Tsujii]{Matsuzaki:2005:PCL:1219840.1219850}
T.~Matsuzaki, Y.~Miyao, and J.~Tsujii.
\newblock Probabilistic cfg with latent annotations.
\newblock In \emph{Proceedings of the 43rd Annual Meeting on Association for
  Computational Linguistics}, ACL '05, pages 75--82, Stroudsburg, PA, USA,
  2005. Association for Computational Linguistics.
\newblock \doi{10.3115/1219840.1219850}.

\bibitem[Mochihashi and Sumita(2007)]{DBLP:conf/nips/MochihashiS07}
D.~Mochihashi and E.~Sumita.
\newblock The infinite markov model.
\newblock In \emph{Advances in Neural Information Processing Systems 20,
  Proceedings of the Twenty-First Annual Conference on Neural Information
  Systems, Vancouver, British Columbia, Canada}, 2007.

\bibitem[Petrov and Klein(2007)]{Petrov-Klein-2007:AAAI}
S.~Petrov and D.~Klein.
\newblock Learning and inference for hierarchically split {PCFG}s.
\newblock In \emph{Proceedings of the Twenty-Second {AAAI} Conference on
  Artificial Intelligence, Vancouver, British Columbia, Canada}, 2007.

\bibitem[Petrov et~al.(2006)Petrov, Barrett, Thibaux, and
  Klein]{petrov2006learning}
S.~Petrov, L.~Barrett, R.~Thibaux, and D.~Klein.
\newblock Learning accurate, compact, and interpretable tree annotation.
\newblock In \emph{Proceedings of the 21st International Conference on
  Computational Linguistics and the 44th annual meeting of the Association for
  Computational Linguistics}, pages 433--440. Association for Computational
  Linguistics, 2006.

\bibitem[Teh(2006)]{teh2006hierarchical}
Y.~W. Teh.
\newblock A hierarchical bayesian language model based on pitman-yor processes.
\newblock In \emph{Proceedings of the 21st International Conference on
  Computational Linguistics and the 44th annual meeting of the Association for
  Computational Linguistics}, pages 985--992. Association for Computational
  Linguistics, 2006.

\bibitem[Thede and Harper(1999)]{Thede:1999:SHM:1034678.1034712}
S.~M. Thede and M.~P. Harper.
\newblock A second-order hidden markov model for part-of-speech tagging.
\newblock In \emph{Proceedings of the 37th Annual Meeting of the Association
  for Computational Linguistics on Computational Linguistics}, ACL '99, pages
  175--182, Stroudsburg, PA, USA, 1999. Association for Computational
  Linguistics.
\newblock ISBN 1-55860-609-3.

\bibitem[Toutanova and Manning(2000)]{toutanova2000enriching}
K.~Toutanova and C.~D. Manning.
\newblock Enriching the knowledge sources used in a maximum entropy
  part-of-speech tagger.
\newblock In \emph{Proceedings of the 2000 Joint SIGDAT conference on Empirical
  methods in natural language processing and very large corpora}, pages 63--70.
  Association for Computational Linguistics, 2000.

\bibitem[Wood et~al.(2009{\natexlab{a}})Wood, Archambeau, Gasthaus, James, and
  Teh]{DBLP:conf/icml/WoodAGJT09}
F.~Wood, C.~Archambeau, J.~Gasthaus, L.~James, and Y.~W. Teh.
\newblock A stochastic memoizer for sequence data.
\newblock In \emph{Proceedings of the 26th Annual International Conference on
  Machine Learning, {ICML} 2009, Montreal, Quebec, Canada, June 14-18, 2009},
  page 142, 2009{\natexlab{a}}.

\bibitem[Wood et~al.(2009{\natexlab{b}})Wood, Archambeau, Gasthaus, James, and
  Teh]{wood2009stochastic}
F.~Wood, C.~Archambeau, J.~Gasthaus, L.~James, and Y.~W. Teh.
\newblock A stochastic memoizer for sequence data.
\newblock In \emph{Proceedings of the 26th Annual International Conference on
  Machine Learning}, pages 1129--1136. ACM, 2009{\natexlab{b}}.

\bibitem[Wood et~al.(2011)Wood, Gasthaus, Archambeau, James, and
  Teh]{wood2011sequence}
F.~Wood, J.~Gasthaus, C.~Archambeau, L.~James, and Y.~W. Teh.
\newblock The sequence memoizer.
\newblock \emph{Communications of the ACM}, 54\penalty0 (2):\penalty0 91--98,
  2011.

\bibitem[Zhu et~al.(1997)Zhu, Byrd, Lu, and Nocedal]{zhu1997algorithm}
C.~Zhu, R.~H. Byrd, P.~Lu, and J.~Nocedal.
\newblock Algorithm 778: L-bfgs-b: Fortran subroutines for large-scale
  bound-constrained optimization.
\newblock \emph{ACM Transactions on Mathematical Software (TOMS)}, 23\penalty0
  (4):\penalty0 550--560, 1997.

\end{thebibliography}
\end{document}